\documentclass[sigconf,preprint]{acmart}
\usepackage{booktabs}
\usepackage{xcolor}
\usepackage{bbding}
\usepackage{graphicx} % for \resizebox

\definecolor{mygreen}{RGB}{34,188,34}
\definecolor{myred}{RGB}{222,34,34}
\newcommand{\cmark}{\textcolor{mygreen}{\CheckmarkBold}}
\newcommand{\xmark}{\textcolor{myred}{\XSolidBrush}}

\setcopyright{acmlicensed}
\copyrightyear{2018}
\acmYear{2018}
\acmDOI{XXXXXXX.XXXXXXX}
\acmConference[Conference acronym 'XX]{Make sure to enter the correct
  conference title from your rights confirmation email}{June 03--05,
  2018}{Woodstock, NY}
\acmISBN{978-1-4503-XXXX-X/2018/06}

\begin{document}
% \usepackage{pifont}

\title{\textsc{Conv-FinRe}: A Conversational and Longitudinal Benchmark for Utility-Grounded Financial Recommendation}

\author{Yan Wang}
\affiliation{%
  \institution{\normalsize The Fin AI}
  \country{\normalsize USA}
}
\email{wy2266336@gmail.com}

\author{Yi Han}
\affiliation{%
  \institution{\normalsize Georgia Institute of Technology}
  \country{\normalsize USA}
}

\author{Lingfei Qian}
\authornote{Corresponding authors.}
\affiliation{%
  \institution{\normalsize The Fin AI}
  \country{\normalsize USA}
}
\email{lingfei.qian@yale.edu}

\author{Yueru He}
\affiliation{%
  \institution{\normalsize Columbia University}
  \country{\normalsize USA}
}

\author{Xueqing Peng}
\authornotemark[1]
\affiliation{%
  \institution{\normalsize The Fin AI}
  \country{\normalsize USA}
}
\email{xueqing.peng@yale.edu}

\author{Dongji Feng}
\authornotemark[1]
\affiliation{%
  \institution{\normalsize California State University}
  \country{\normalsize USA}
}
\email{dfeng@csumb.edu}

\author{Zhuohan Xie}
\affiliation{%
  \institution{\normalsize MBZUAI}
  % \city{Haidian Qu}
  % \state{Beijing Shi}
  \country{\normalsize UAE}}

\author{Vincent Jim Zhang}
\affiliation{%
  \institution{\normalsize The Fin AI}
  \country{\normalsize USA}
}

\author{Rosie Guo}
\affiliation{%
  \institution{\normalsize The Fin AI}
  \country{\normalsize USA}
}

\author{Fengran Mo}
\affiliation{%
  \institution{\normalsize University of Montreal}
  % \city{Hekla}
  \country{\normalsize Canada}}
  
\author{Jimin Huang}
\affiliation{%
   \institution{The University of Manchester}
   \city{Manchester}
   \country{United Kingdom}
}
\affiliation{%
  \institution{\normalsize The Fin AI}
  \country{\normalsize USA}
}

\author{Yankai Chen}
\authornotemark[1]
\affiliation{%
  \institution{\normalsize McGill University \\ MBZUAI}
  \country{\normalsize Canada}}
\email{yankaichan3@gmail.com}

\author{Xue (Steve) Liu}
\affiliation{%
  \institution{\normalsize McGill University \\ MBZUAI}
  \country{\normalsize Canada}}
  
\author{Jian-Yun Nie}
\affiliation{%
  \institution{\normalsize University of Montreal}
  % \city{New York}
  \country{\normalsize Canada}}

\renewcommand{\shortauthors}{Yan et al.}

\begin{abstract}
Most recommendation benchmarks evaluate how well a model imitates user behavior. In financial advisory, however, observed actions can be noisy or short-sighted under market volatility and may conflict with a user’s long-term goals. Treating \emph{what users chose} as the sole ground truth, therefore, conflates behavioral imitation with decision quality.
We introduce \textsc{Conv-FinRe}, a conversational and longitudinal benchmark for stock recommendation that evaluates LLMs beyond behavior matching. Given an onboarding interview, step-wise market context, and advisory dialogues, models must generate rankings over a fixed investment horizon. Crucially, \textsc{Conv-FinRe} provides multi-view references that distinguish descriptive behavior from normative utility grounded in investor-specific risk preferences, enabling diagnosis of whether an LLM follows rational analysis, mimics user noise, or is driven by market momentum. We build the benchmark from real market data and human decision trajectories, instantiate controlled advisory conversations, and evaluate a suite of state-of-the-art LLMs. Results reveal a persistent tension between rational decision quality and behavioral alignment: models that perform well on utility-based ranking often fail to match user choices, whereas behaviorally aligned models can overfit short-term noise. The dataset is publicly released on Hugging Face\footnote{\url{https://huggingface.co/collections/TheFinAI/conv-finre}}, and the codebase is available on GitHub\footnote{\url{https://github.com/The-FinAI/Conv-FinRe}}.

  % While Large Language Models (LLMs) show promise in personalized financial advisory, current evaluation paradigms often conflate behavioral imitation with genuine decision utility. We present \textsc{Conv-FinRe}, a longitudinal benchmark that redefines stock recommendation as a multi-view alignment task. By integrating inverse optimization, our framework reconstructs latent investor profiles from interaction traces, providing a principled mechanism to audit how models navigate conflicting financial objectives over a temporal horizon. Our analysis of leading LLMs exposes a critical dichotomy between normative rationality and descriptive alignment. We find that while high-capacity models demonstrate technical proficiency in utility-based ranking, they lack the "advisory intelligence" to disentangle structural risk from market volatility. Conversely, domain-tuned models often prioritize superficial mimicry, treating stochastic user actions as definitive preferences. By categorizing model behaviors into distinct archetypes, this work reveals a systemic bottleneck in reconciling objective financial principles with subjective investor intent, establishing a new rigor for developing robust, interactive financial agents.
\end{abstract}






\maketitle

\section{Introduction}
Large language models (LLMs) have achieved remarkable progress across diverse application domains, demonstrating strong performance in language understanding~\cite{yu2025benchmarking,peng2025multifinben}, reasoning~\cite{qian2025fino1transferabilityreasoningenhancedllms,zhao2024docmathevalevaluatingmathreasoning}, and structured problem-solving~\cite{wang2026fintaggingbenchmarkingllmsextracting,wang2025finauditingfinancialtaxonomystructuredmultidocument}.
These capabilities have motivated their adoption as assistants for decision-making and recommendation tasks~\cite{li2025investorbench,liu2025benchmarking,di2023evaluating,dai2023uncovering,lyu2024llm}. In most recommendation benchmarks, personalization is primarily measured by behavioral imitation: a recommendation is deemed correct if it matches what a user would click, rate, or choose~\cite{chen2023reasoner,sah2025faireval,lee2026cereal,liang2024llm}. This behavior-centric paradigm is effective in many consumer domains, where feedback is a reliable proxy for utility.

\begin{table}[t]
\caption{Comparison of representative user-centric recommendation benchmarks.
\textit{Dynamic} indicates time-conditioned signals; \textit{Utility} refers to relevance grounded in a user-dependent decision utility;
\textit{Multi-view} indicates multiple, potentially conflicting ranking views; \textit{Dialogue} denotes a conversational interactive setting.}
\label{tab:relatedwork-compare}
\centering
\resizebox{\linewidth}{!}{%
\begin{tabular}{lccccc}
\toprule
\textbf{Work} &
\textbf{Domain} &
\textbf{Dynamic} &
\textbf{Utility} &
\textbf{Multi-view} &
\textbf{Dialogue} \\
\midrule
REASONER~\cite{chen2023reasoner} & Video & \xmark & \xmark & \xmark & \xmark  \\
FairEval~\cite{sah2025faireval} & Music & \xmark & \xmark & \xmark & \xmark  \\
PerFairX~\cite{sah2025perfairx} & Movie\&Music  & \xmark & \xmark & \xmark & \xmark \\
CEREAL~\cite{lee2026cereal} & Movie & \xmark & \xmark & \xmark & \cmark \\
LLM-REDIAL~\cite{liang2024llm} & Movie & \xmark & \xmark & \xmark & \cmark \\
RecBench~\cite{liu2025benchmarking} & General  & \xmark & \xmark & \xmark & \xmark \\
FAR-Trans~\cite{sanz2024far} & Finance  & \cmark & \xmark & \xmark & \xmark \\
\midrule
\textbf{Ours} & Finance & \cmark & \cmark & \cmark & \cmark  \\
\bottomrule
\end{tabular}%
}
\end{table}

Financial recommendation is different. Investor actions are often affected by short-term market noise, emotions, and shifting constraints, and may deviate from stable risk tolerance or long-term objectives~\cite{arran2023behavioral,mehraj2025psychological}. As a result, matching historical choices alone cannot tell whether an advisor is providing \emph{good} financial guidance. A faithful mimic of noisy actions may be misaligned with the user’s underlying goals, while a purely rational advisor may ignore user intent and preferences.

Existing benchmarks, as summarized in Table~\ref{tab:relatedwork-compare}, therefore struggle with three issues: \emph{behavior-as-truth}, \emph{utility blindness}, and \emph{single-view evaluation}. Most user-centric benchmarks rely on relevance signals (clicks/ratings)~\cite{liu2025benchmarking,chen2023reasoner} without utility grounding, while finance datasets often emphasize prediction or trading objectives rather than user-specific decision quality~\cite{sanz2024far}. Consequently, they cannot diagnose whether an LLM advisor is reasoning about risk-sensitive utility, blindly chasing market trends, or simply overfitting user noise.

% However, as summarized in Table~\ref{tab:relatedwork-compare}, existing benchmarks are primarily characterized by \emph{relevance-based personalization}~\cite{chen2023reasoner,sah2025faireval,sah2025perfairx,lee2026cereal}, \emph{prediction-induced ranking}~\cite{sanz2024far}, and \emph{static evaluation}~\cite{liang2024llm,liu2025benchmarking}. Most recommendation benchmarks reduce "user-centricity" to behavioral similarity (e.g., clicks or ratings)~\cite{liu2025benchmarking,chen2023reasoner}, which fails to capture the underlying financial utility. In the financial domain, benchmarks typically focus on price forecasting, treating recommendation as a by-product of prediction rather than a first-class objective that must reconcile market dynamics with shifting investor psychology~\cite{sanz2024far}. Consequently, it remains unclear whether LLM advisors can genuinely discover and optimize latent user objectives in dynamic, multi-view, and interactive settings.

% As a result, it remains unclear whether LLM advisors can discover and follow user-aligned objectives under dynamic, multi-objective, and interactive decision settings such as personalized stock recommendations. Addressing this gap requires benchmarks that jointly model longitudinal preference discovery, explicit decision utility, and conversational interaction, requirements that are not simultaneously satisfied by existing work. ion-making settings, such as personalized stock recommendations?

To address this gap, we introduce \textsc{Conv-FinRe}, the first conversational and longitudinal benchmark that formulates financial recommendation as a \emph{multi-view alignment} problem.
Rather than evaluating whether an LLM simply matches user choices, the benchmark assesses model rankings against four complementary reference views, such as user choice ($y_{user}$), rational utility ($y_{util}$), market momentum ($y_{mom}$), and risk sensitivity ($y_{safe}$), enabling diagnosis of whether a model relies on rational analysis, behavioral imitation, or short-term market signals. To support such evaluation, user-specific risk preferences are inferred from longitudinal decision trajectories via inverse optimization and used to construct utility- and risk-based reference rankings, without exposing the latent utility function to the model.
Operationally, \textsc{Conv-FinRe} instantiates the task through onboarding interviews and step-wise advisory dialogues, where an LLM must reconcile competing advisory principles over time.

% To address this gap, we introduce \textsc{Conv-FinRe}, the first interactive benchmark for personality-grounded multi-view longitudinal stock recommendation. Unlike traditional benchmarks, \textsc{Conv-FinRe} simulates a complete advisory lifecycle: starting from an onboarding interview to elicit initial traits, followed by sequential decision-making over a 30-day market horizon. Our framework introduces three key innovations: (1) we employ inverse optimization to infer user-specific risk parameters from behavioral traces, establishing four complementary reference views, such as User Choice, Rational Utility, Market Momentum, and Risk Sensitivity, to enable multi-dimensional alignment analysis. (2) Unlike single-objective rankers, our benchmark requires LLMs to act as mediators that reconcile conflicting signals from a "Panel of Experts", testing their ability to handle real-world advisory trade-offs. (3) By isolating the impact of conversational history, we categorize LLMs into three archetypes, such as Adaptive Advisors, Transaction-driven Analysts, and Behavioral Overfitters, to diagnose whether a model captures long-term preference logic or merely mimics short-term noise.

We evaluate a diverse set of state-of-the-art LLMs under \textsc{Conv-FinRe} and reveal a fundamental tension between rational decision quality and behavioral alignment. While some models achieve strong utility-based rankings, they often conflate long-term risk with short-term market momentum; conversely, domain-specialized models tend to overfit noisy user actions, mistaking transient behavior for stable preferences. Together, these results motivate \textsc{Conv-FinRe} as a benchmark for multi-view, utility-grounded evaluation.

We make the following contributions:
(1) We introduce \textsc{Conv-FinRe}, a conversational and longitudinal benchmark for stock recommendation that evaluates LLMs beyond behavioral imitation by grounding assessment in investor-specific utility.
(2) We formulate financial recommendation evaluation as a multi-view alignment problem and provide a diagnostic framework, supported by inverse optimization, that disentangles behavioral alignment from rational decision quality.
(3) We conduct a systematic evaluation of state-of-the-art LLMs under \textsc{Conv-FinRe}, identifying distinct advisory behavior patterns under competing market signals and user noise.

\section{Related Works}

Personalized recommendation benchmarks are well studied in consumer domains such as e-commerce and media, where personalization is typically modeled from interaction histories or coarse user traits and supervision relies on a single relevance signal like ratings or clicks \cite{chen2023reasoner,yang2022personality}. Recent work enriches this setting by introducing structured explanations \cite{chen2023reasoner} and by evaluating LLMs either as representation enhancers or as end-to-end recommenders under point-wise, pair-wise, or list-wise protocols \cite{lyu2024llm,dai2023uncovering,di2023evaluating,liu2025benchmarking}, with further analysis of fairness, bias, and sequential alignment \cite{sah2025faireval,liao2023llara}. In contrast, personalized stock recommendation poses additional challenges due to non-stationary assets and investor objectives constrained by risk tolerance and return–risk trade-offs \cite{takayanagi2023personalized,sanz2024far}, with evidence that risk preferences evolve and must be inferred from behavior rather than static profiles \cite{capponi2020risk}. While conversational and LLM-based financial advisors enable iterative preference elicitation \cite{sun2018conversational,gao2021advances,sharma2021stockbabble,takayanagi2025finpersona,takayanagi2025generative,oehler2024does}, existing benchmarks rarely evaluate recommendation quality using investor-specific utility as the core supervision signal, limiting their ability to assess true decision alignment.

\section{\textsc{Conv-FinRe}}

\begin{figure}[h]
  \centering
  \includegraphics[width=\linewidth]{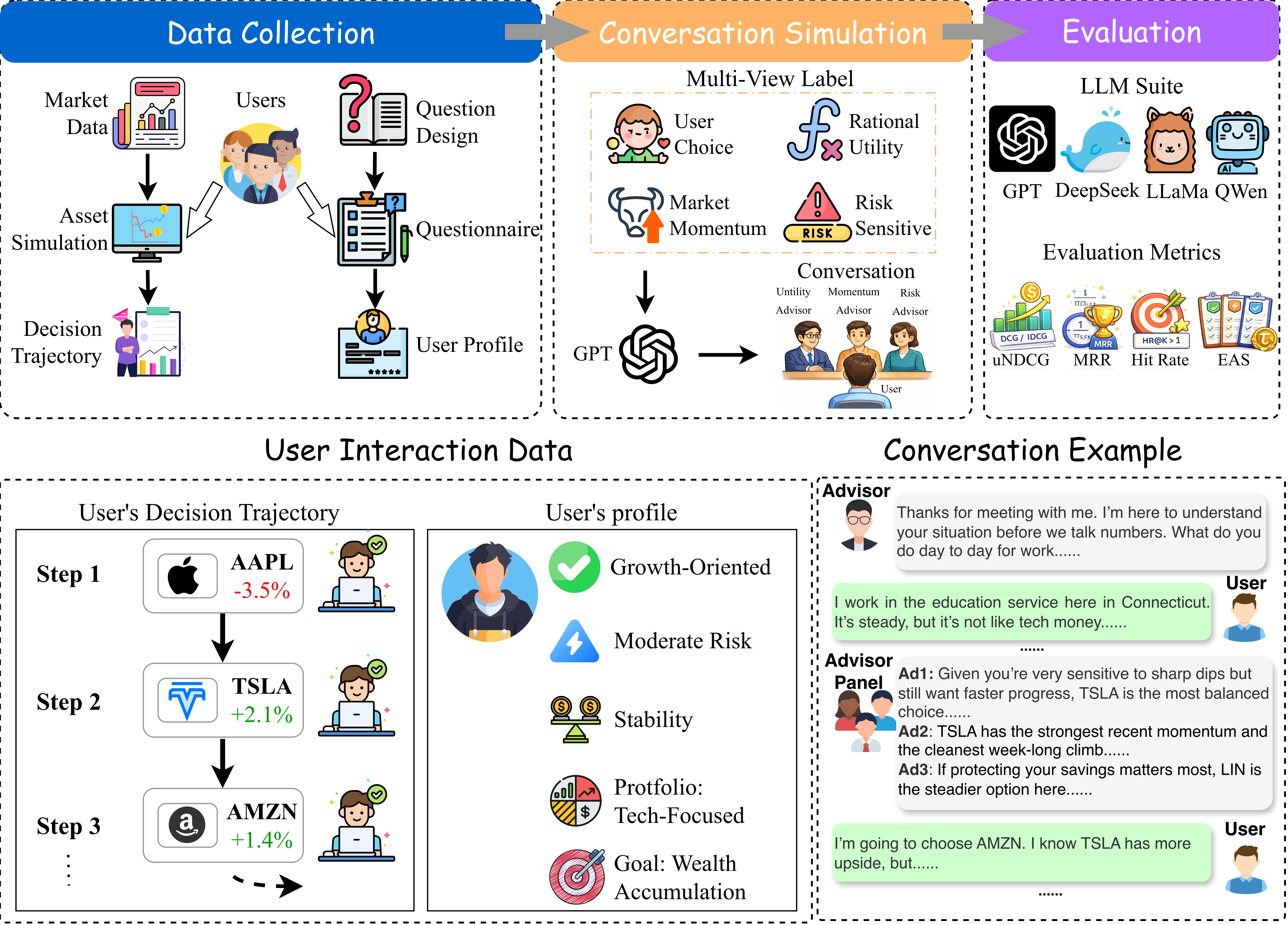}
  \caption{Overview of \textsc{Conv-FinRe} benchmark.}\label{fig:framework}
\end{figure}

Figure~\ref{fig:framework} illustrates the overall pipeline of Conv-FinRe, from data collection and user profiling to multi-view conversation simulation and evaluation.
The framework models longitudinal advisory interactions by integrating market signals, inferred user preferences, and competing expert recommendations, enabling fine-grained analysis of LLM alignment in personalized financial decision-making.

\subsection{Task Formulation}

We define the \textbf{Multi-view Longitudinal Stock Recommendation} task, which simulates iterative interactions between a personalized investment advisor and a user over a fixed investment horizon $T$. Unlike conventional recommendation benchmarks that rely on a single "gold-standard" label, our task evaluates LLM alignment across four complementary reference views: \textbf{User Choice ($y_{user}$)}, representing the empirical selections made by the human participant; \textbf{Rational Utility ($y_{util}$)}, an idealized ranking derived from a calibrated utility function that represents the theoretically optimal balance between return and risk; \textbf{Market Momentum ($y_{mom}$)}, a profit-oriented ranking based purely on recent cumulative returns; and \textbf{Risk Sensitivity ($y_{safe}$)}, a conservative ranking that isolates the user's specific risk-avoidance component by penalizing volatility and downside risk according to their inferred sensitivity.

In this setting, the LLM acts as a \textbf{Personalized Investment Advisor} whose objective is not explicitly specified. Instead, the model must infer the user’s latent financial preferences over time and reconcile conflicting advisory principles. At each decision step, the model is presented with the current market context, historical interaction trajectory, and recommendations from a panel of three specialized advisors grounded in $y_{util}$, $y_{mom}$, and $y_{safe}$. The core challenge for the LLM is to synthesize these heterogeneous signals to produce a final ranking that reflects its inferred understanding of the user’s underlying financial objectives. This multi-view design enables us to diagnose whether an LLM's misalignment with actual user behavior stems from an over-reliance on market momentum or a miscalculation of the user's specific risk thresholds.

Formally, for the user $i$ at step $t$, given a candidate stock set $\mathcal{S}_t$ and decision context $\mathcal{I}^i_t$, the recommendation process is defined as:

\begin{equation} 
    \pi_{i,t} = f_\theta (\mathcal{I}^i_t, \mathcal{S}_t), 
\end{equation}
where $f_\theta$ denotes the evaluated LLM-based advisor, $\pi_{i,t}$ is the reranked list from the $\mathcal{S}_t$. Specifically, the decision context $\mathcal{I}^i_t$ is shown below:
\begin{equation}
    \mathcal{I}^i_t = \left( P^i, \mathcal{H}^i_{1:t-1}, \mathcal{M}t \right),
\end{equation}
where $P^i$ denotes the onboarding dialogue, a multi-turn introductory interaction used to elicit the user $i$'s background, financial goals, and initial risk tolerance. $\mathcal{M}_t = \{ \mathbf{v}_{s,t}, \mathbf{x}_{s,t} \}_{s \in \mathcal{S}_t}$ represents the current market state. Here, $\mathbf{v}_{s,t} = (\mu_{s}, \sigma^2_{s}, \text{Drawdown}_{s})$ denotes the vector of raw performance metrics used for preference grounding, while $\mathbf{x}_{s,t}$ comprises the verbalized market signals (e.g., price trends, percentage returns, and volatility) presented to the LLM advisor. The term $\mathcal{H}^i_{1:t-1}$ encapsulates the longitudinal interaction trajectory, consisting of multi-turn dialogues between the user and the three specialized advisors ({Rational Utility}, {Market Momentum}, and {Risk Sensitivity}). This history provides the model with both the user’s previous decision patterns and the conflicting advisory signals previously encountered.

\subsection{Latent Preference Grounding via Inverse Optimization}

The task itself does not assume access to the user’s true utility function. 
Instead, to enable principled analysis of advisory behaviors, we construct a latent preference signal that serves as a reference representation of each user’s underlying risk attitude.
This preference signal is used to characterize advisory objectives within the benchmark, rather than being exposed to the model.

We assume that user $i$’s decision-making process is governed by a latent utility function $U_{i,t}^{(s)}$, which balances expected return against volatility and downside risk~\cite{rubinstein2002markowitz,magdon2004maximum,tversky1992advances}. For user $i$ and stock $s \in \mathcal{S}_t$ at step $t$, the utility is defined as:

\begin{equation}
U_{i,t}^{(s)} = \tilde{\mu}_{s,t} - \lambda_i \tilde{\sigma}_{s,t}^2 - \gamma_i \tilde{\text{Drawdown}}_{s,t},
\end{equation}
where $\tilde{\mu}_{s,t}$, $\tilde{\sigma}_{s,t}^2$, and $\tilde{\text{Drawdown}}_{s,t}$ denote the cross-sectionally standardized mean return, variance, and maximum drawdown of stock $s$ over a 7-day window preceding step $t$. $(\lambda_i, \gamma_i)$ are the user-specific parameters, which are assumed to be time-invariant and capture the user’s sensitivity to volatility and downside risk.

We estimate $(\lambda_i, \gamma_i)$ via \emph{Inverse Optimization}~\cite{bertsimas2012inverse,bertsimas2015data} using the user’s longitudinal behavioral trajectory $H^i_{1:T}$. Assuming a rational choice model with Gumbel-distributed noise, the probability that user $i$ selects stock $s^*$ at step $t$ follows a Multinomial Logit model~\cite{so1995multinomial}:

\begin{equation}
P(s^* \mid \lambda_i, \gamma_i, \mathcal{M}_t) 
= \frac{\exp\left(U_{i,t}^{(s^*)}\right)}
{\sum_{s \in \mathcal{S}_t} \exp\left(U_{i,t}^{(s)}\right)},
\end{equation}
% where $\mathcal{S}_t$ denotes the candidate stock set at step $t$ and $\mathcal{M}_t$ represents the market state.

The global parameters are obtained by minimizing the Regularized Negative Log-Likelihood~\cite{bertero2006regularization,mcfadden1972conditional} over the interaction horizon:

\begin{equation}
\begin{aligned}
\mathcal{L}_i(\lambda_i, \gamma_i) =\;&
-\sum_{t=1}^{T}
\log P\!\left(
s^*_{i,t} \mid \lambda_i, \gamma_i, \mathcal{M}_t
\right) \\
&+ \alpha \left\| (\lambda_i, \gamma_i) \right\|_2^2 ,
\end{aligned}
\end{equation}

\begin{equation}
(\lambda_i, \gamma_i) =
\arg\min_{\lambda_i, \gamma_i} \mathcal{L}_i .
\end{equation}
where $s^*_{i,t}$ is the stock actually chosen by user $i$ at step $t$, and $\alpha$ controls the strength of regularization.

Once $(\lambda_i, \gamma_i)$ are estimated, we construct the reference views for evaluation. Specifically, the \textbf{Rational Utility} view ($y_{util}$) is ranked by the full utility $U_{i,t}^{(s)}$, while the \textbf{Risk Sensitivity} view ($y_{safe}$) is ranked by the personalized risk penalty term: $R_{i,t}^{(s)} = \lambda_i \tilde{\sigma}_{s,t}^2 + \gamma_i \tilde{\text{Drawdown}}_{s,t}$.

\subsection{Data Collection and Conversation Simulation}
\subsubsection{Data Collection}
\paragraph{Market Data} We construct a compact stock universe from S\&P~500 constituents using stratified sampling to ensure coverage of all eleven GICS sectors and balanced exposure to systematic risk.
Candidate stocks are grouped by market beta, computed from five-year monthly returns, into low ($\beta<1$), moderate ($\beta\approx1$), and high ($\beta>1$) risk regimes, with approximately equal representation from each group.
The resulting universe comprises ten representative stocks (Table~\ref{tab:stock_universe}), enabling controlled longitudinal evaluation.

For each stock, we collect daily and intraday price data over a 30-day horizon (Aug.~6--Sep.~17,~2025) via the Yahoo Finance API\footnote{\url{https://pypi.org/project/yfinance/}}, which defines the market state in the simulation environment.

\begin{table}[t]
\centering
\caption{Summary of the stock universe used in \textsc{Conv-FinRe}.
Stocks are grouped by volatility tier based on market beta to ensure balanced exposure to systematic risk.}
\label{tab:stock_universe}
\resizebox{\linewidth}{!}{%
\begin{tabular}{l l l c l}
\hline
\textbf{Ticker} & \textbf{Company} & \textbf{GICS Sector} & \textbf{Beta} & \textbf{Market Cap} \\
\hline
\multicolumn{5}{l}{\textbf{Low Volatility ($\beta < 1$)}} \\
\hline
PG   & Procter \& Gamble           & Consumer Staples        & 0.36 & \$353B \\
% DUK  & Duke Energy Corporation     & Utilities               & 0.38 & \$94B \\
MRK  & Merck \& Co., Inc.          & Health Care             & 0.38 & \$274B \\
VZ   & Verizon Communications Inc. & Communication Services  & 0.36 & \$188B \\
\hline
\multicolumn{5}{l}{\textbf{Moderate Volatility ($\beta \approx 1$)}} \\
\hline
LIN  & Linde plc                   & Materials               & 0.95 & \$213B \\
XOM  & Exxon Mobil Corporation     & Energy                  & 0.95 & \$596B \\
\hline
\multicolumn{5}{l}{\textbf{High Volatility ($\beta > 1$)}} \\
\hline
JPM  & JPMorgan Chase \& Co.       & Financials              & 1.13 & \$841B \\
AMZN & Amazon.com, Inc.            & Information Technology  & 1.31 & \$2.6T \\
MMM  & 3M Company                  & Industrials             & 1.10 & \$81B \\
SPG  & Simon Property Group, Inc.  & Real Estate             & 1.53 & \$62B \\
TSLA & Tesla, Inc.                 & Consumer Discretionary  & 2.07 & \$1.4T \\
\hline
\end{tabular}%
}
\end{table}

\paragraph{User Interaction Data}
User interaction data are collected through a two-stage protocol.
First, we obtain static user profiles from 10 participants via a structured questionnaire\footnote{\url{https://forms.gle/g7GnwqByq7mCoJgTA}} capturing investor demographics, financial capacity, investment experience, and risk attitudes.
The questionnaire design follows regulatory suitability guidelines (MiFID II Article~25\footnote{\url{https://eur-lex.europa.eu/eli/dir/2014/65/oj/}}; FINRA Rule~2111\footnote{\url{ https://www.finra.org/rules-guidance/rulebooks/finra-rules/2111}}) and is informed by Betterment's approach\footnote{\url{https://d-nb.info/116404222X}} on investor risk preferences and financial decision-making; and industry examples, such as JPMorgan Financial Health Check\footnote{\url{https://am.jpmorgan.com/content/dam/jpm-am-aem/asiapacific/hk/en/literature/account-forms/healthcheck_corporate.pdf}}, Charles Schwab Investor Risk Profile\footnote{\url{https://www.studocu.vn/vn/document/royal-melbourne-institute-of-technology-vietnam/international-trade/charles-schwab-investor-risk-profile/100562992}}, Vanguard Investor Financial Profile\footnote{\url{https://sustainableinvest.com/wp-content/uploads/Investor-Financial-Profile-Questionnaire.pdf}}, and Financial Group Plan\footnote{\url{https://financialgroup.com/risk-profile-bq}}.

Second, we collect longitudinal decision trajectories using a custom asset simulation tool.
Participants interact with a fixed universe of ten stocks over a 30-day horizon, observing daily and intraday price movements.
At each step, users make incremental buy decisions and receive portfolio-level feedback, including realized returns and volatility, which is logged together with their actions to form a temporally ordered interaction trace.
The simulation tool is publicly released for reproducibility\footnote{\url{https://huggingface.co/spaces/TheFinAI/LetYourProfitsRun}}.

\subsubsection{Conversation Simulation}
\paragraph{Conversation Generation} Building on the collected user interaction data, we construct a structured conversation simulation to instantiate the proposed longitudinal advisory task in a language-based setting. Rather than collecting free-form dialogues from participants, we transform each user’s observed profile and behavioral trajectory into a coherent multi-turn advisory conversation, enabling controlled and reproducible evaluation of LLMs. Each user trajectory is organized into two phases: an \emph{onboarding interview} and a longitudinal advisory dialogue over a fixed horizon $T$.

The onboarding phase verbalizes the static user profile obtained from the questionnaire. Using the user’s survey responses as grounding signals, we generate a four-turn advisor--user dialogue that captures the user’s financial background, constraints, investment goals, and emotional reactions to risk. The generated language is constrained to match the user’s reported financial literacy, ensuring that preference signals are conveyed implicitly through natural expression rather than explicit financial terminology. This onboarding dialogue serves as the conversational realization of $P^i$ in the decision context $\mathcal{I}^i_t$.

Following onboarding, the longitudinal advisory phase reflects the user’s sequential decision-making behavior observed in the asset simulation. At each step $t$, the conversation conditions on the historical interaction trajectory $\mathcal{H}^i_{1:t-1}$ and the current market state $\mathcal{M}_t$, represented by a 7-day market snapshot derived from the simulation environment. A panel of three specialized advisors provides recommendations based on the heterogeneous principles defined in our multi-view framework: \textbf{Rational-Utility}, \textbf{Market-Momentum}, and \textbf{Risk-Sensitivity} strategies. The user then finalizes a choice, potentially deviating from advisor suggestions and providing a subjective justification consistent with their observed behavior in the simulation. All advisor messages and user responses are appended to the conversation history, yielding a temporally ordered dialogue aligned with the underlying interaction trace.
Overall, as shown in Table~\ref{tab:conv_stats}, the benchmark contains 10 users and 230 prefix-conditioned instances. Dialogue context grows from 4 to 26 turns (15 on average), yielding 270 unique turns that expand to 3,450 prefix-conditioned turns. Each instance contains 1,818–7,252 tokens (4,320.2 on average), highlighting the substantial longitudinal context complexity required for evaluation.

\begin{table}[t]
\centering
\caption{Structural statistics of Conv-FinRe benchmark. Token counts are computed using the \texttt{cl100k\_base} tokenizer.}
\label{tab:conv_stats}
\resizebox{0.9\linewidth}{!}{
\begin{tabular}{l c l c}
\hline
\multicolumn{2}{c}{\textbf{Scale}} &
\multicolumn{2}{c}{\textbf{Dialogue}} \\
\hline
Users & 10 &
Min Turns / Instance & 4 \\
Steps / User & 23 &
Max Turns / Instance & 26 \\
Total Instances & 230 &
Avg Turns / Instance & 15 \\

\midrule
\multicolumn{2}{c}{\textbf{Turn Accounting}} &
\multicolumn{2}{c}{\textbf{Token Status}} \\
\hline
Unique Turns & 270 &
Min Tokens / Instance & 1,818 \\
Prefix-Expanded Turns & 3,450 &
Max Tokens / Instance & 7,252 \\
& &
Avg Tokens / Instance & 4,320.2 \\
\hline
\end{tabular}
}
\end{table}

% \begin{table}[t]
% \small
% \centering
% \caption{Statistics of the Conv-FinRe benchmark.}
% \label{tab:conv_stats}
% \begin{tabular}{l c}
% \hline
% \textbf{Statistic} & \textbf{Value} \\
% \hline
% Users & 10 \\
% Horizon (T) & 23 \\
% Total Instances & 230 \\
% Avg Dialog. Turns / Instance & XXX \\
% Total Tokens & XXX \\
% Avg Tokens / Conv & XXX \\
% \hline
% \end{tabular}
% \end{table}

\paragraph{Conversation Quality Validation} We validate the quality of the simulated conversations from both a preference-grounding and a conversational realism perspective.

\textbf{User Preference Consistency Validation:} For each user $i$, the inferred latent parameters $(\lambda_i, \gamma_i)$ represent the user’s sensitivity to volatility and downside risk. To assess whether these parameters meaningfully capture the user’s true investment psychology, financial experts translate each parameter pair into a concise natural-language summary of the user’s risk tolerance. Users are then asked to rate, on a 0--9 Likert scale (0 = not at all accurate, 9 = perfectly accurate), how well the summary reflects their own reasoning and emotions when making stock selection decisions. Across users, the summaries receive a high average agreement score of \textbf{7.8}, with low inter-user variance, indicating that the inferred preferences are largely consistent with users' decision-making behavior.

\textbf{Conversational Plausibility Validation:} We further assess the realism of the simulated advisory dialogues. To avoid over-counting highly correlated interactions within the same user trajectory, we adopt a user-level sampling strategy. Specifically, for each of the 10 users, we randomly sample one decision step from the 23-step longitudinal conversation, resulting in 10 representative dialogue instances. Each sampled conversation is independently evaluated by three domain experts with backgrounds in finance and conversational systems. Experts rate each dialogue on a 0--9 Likert scale along four dimensions: role consistency, linguistic naturalness, behavioral plausibility of user responses, and cross-turn coherence. The final plausibility score for each conversation is computed as the average across the four dimensions and across experts. Overall, the simulated dialogues achieve a mean plausibility score of \textbf{8.1}, indicating that the generated conversations closely resemble realistic financial advisory interactions rather than scripted or artificial exchanges.

\subsection{Evaluation Metrics}

% \paragraph{Utility-based NDCG (uNDCG)}

\noindent\textbf{Utility-based NDCG (uNDCG):}
We first evaluate whether model-generated rankings align with the user's latent, utility-grounded preference structure.
For user $i$ at step $t$, we compute uNDCG using the calibrated utility $U_{i,t}^{(s)}$ as relevance.
Given a ranking $\pi_{i,t}$ over candidate set $\mathcal{S}_t$, the discounted cumulative gain is:
\begin{equation}
\text{DCG}_{i,t}
= \sum_{k=1}^{|\mathcal{S}_t|}
\frac{U_{i,t}^{(\pi_{i,t}[k])}}{\log_2(k+1)} ,
\end{equation}
and the utility-based NDCG is defined as:
\begin{equation}
\text{uNDCG}_{i,t}
= \frac{\text{DCG}_{i,t}}{\text{IDCG}_{i,t}},
\end{equation}
where $\text{IDCG}_{i,t}$ is computed from the utility-optimal ranking. 

% uNDCG measures global ranking quality under risk-sensitive trade-offs.

% \paragraph{MRR and Hit Rate}

\noindent\textbf{MRR and Hit Rate:}
To assess recovery of the user's observed choice, let $s^*_{i,t}$ denote the stock selected by user $i$ at step $t$, and let $\pi_{i,t}(s^*_{i,t})$ be its position in $\pi_{i,t}$.
The reciprocal rank is
\begin{equation}
\text{RR}_{i,t}
= \frac{1}{\pi_{i,t}(s^*_{i,t})},
\end{equation}
with Mean Reciprocal Rank (MRR) obtained by averaging across users and steps.
We additionally report Hit Rate at top-$K$:
\begin{equation}
\text{HR@}K_{i,t}
=
\mathbb{I}\!\left[\pi_{i,t}(s^*_{i,t}) \le K\right],
\end{equation}
and focus on $K \in \{1,3\}$, where $\mathbb{I}[\cdot]$ denotes the indicator function.

% \paragraph{Expert Alignment Score (EAS)}

\noindent\textbf{Expert Alignment Score (EAS):}
To analyze how models resolve competing advisory principles, we measure alignment with three expert rankings: \textbf{Rational Utility}, \textbf{Market Momentum}, and \textbf{Risk Sensitivity}.
For model $m$, we compute step-wise alignment using Kendall’s $\tau$:
\begin{equation}
\text{EAS}^{(e)}_{i,t}(m)
= \tau\!\left(\pi^{m}_{i,t}, \pi^{e}_{i,t}\right),
\end{equation}
where $e$ denotes the expert type.
Final scores are obtained by averaging over all users and steps:
\begin{equation}
\overline{\text{EAS}}^{(e)}(m)
= \frac{1}{NT}\sum_{i=1}^{N}\sum_{t=1}^{T}
\text{EAS}^{(e)}_{i,t}(m).
\end{equation}

\subsection{Evaluation Models}

Our goal is to assess the capabilities and limitations of contemporary LLMs in \emph{conversational, personality-grounded longitudinal stock recommendation} under the \textsc{Conv-FinRe} benchmark.
To this end, we evaluate a diverse set of state-of-the-art LLMs, covering both proprietary and open-source families.
Specifically, we include two closed-source general-purpose models, \textbf{GPT-5.2}~\cite{singh2025openaigpt5card} and \textbf{GPT-4o}~\cite{hurst2024gpt}. %as well as \textbf{Gemini-3}.
We further evaluate a range of open-source general models with strong reasoning and instruction-following capabilities, including \textbf{DeepSeek-V3.2}~\cite{deepseekai2025deepseekv32}, \textbf{Qwen3-235B-A22B-Instruct}~\cite{qwen3technicalreport}, \textbf{Qwen2.5-72B-Instruct}~\cite{qwen2}, and \textbf{Llama-3.3-70B-Instruct}~\cite{grattafiori2024llama}, and one financial domain conversational model: \textbf{Llama3-XuanYuan3-70B-Chat}\footnote{\url{https://huggingface.co/Duxiaoman-DI/Llama3-XuanYuan3-70B-Chat}}.
All models are evaluated using the LM Evaluation Harness~\cite{eval-harness} under a unified interface.
Proprietary models are accessed via official APIs, while open-source models are executed locally.
Across all experiments, we standardize the maximum input context length to 8,192 tokens and constrain the maximum generation length to 126 tokens, ensuring fair and consistent comparison across models.

\section{Experiments and Results}

\subsection{Overall Performance}

Table~\ref{tab:overall-performance} shows that most models achieve high uNDCG scores (0.92–0.97), indicating a strong baseline for ranking assets according to the Rational Utility. However, high uNDCG does not always translate into better recovery of the User Choice. While Llama-3.3-70B-Instruct leads in uNDCG (0.97), it shows lower Hit Rates, suggesting it prioritizes an "idealized" rational recommendation that balances long-term risk and return.

\begin{table}[t]
\centering
\caption{Overall performance on the longitudinal stock advisory task. The Random baseline is computed by averaging over 1,000 uniform random permutations per instance, serving as a sanity-check lower bound.}
\label{tab:overall-performance}
\resizebox{\linewidth}{!}{%
\begin{tabular}{lcccc}
\hline
Model & uNDCG$\uparrow$ & MRR$\uparrow$ & HR@1$\uparrow$ & HR@3$\uparrow$ \\
\hline
Random & $0.73 \pm 0.00$ & $0.29 \pm 0.01$ & $0.10 \pm 0.01$ & $0.30 \pm 0.01$ \\
GPT-5.2 & $0.94 \pm 0.03$ & $0.46 \pm 0.02$ & $0.29 \pm 0.03$ & $0.51 \pm 0.03$ \\
GPT-4o & $0.94 \pm 0.00$ & $0.56 \pm 0.03$ & $0.42 \pm 0.03$ & $0.60 \pm 0.03$ \\
% Gemini-3 & -- & -- & -- & -- \\
DeepSeek-V3.2 & $0.92 \pm 0.00$ & $0.51 \pm 0.03$ & $0.37 \pm 0.03$ & $0.55 \pm 0.03$ \\
Qwen3-235B-A22B-Instruct & $0.94 \pm 0.00$ & $0.47 \pm 0.02$ & $0.30 \pm 0.03$ & $0.52 \pm 0.03$ \\
Qwen2.5-72B-Instruct & $0.92 \pm 0.01$ & $0.63 \pm 0.03$ & $0.50 \pm 0.03$ & $0.69 \pm 0.03$ \\
Llama-3.3-70B-Instruct & $0.97 \pm 0.00$ & $0.52 \pm 0.03$ & $0.36 \pm 0.03$ & $0.59 \pm 0.03$ \\
Llama3-XuanYuan3-70B-Chat & $0.92 \pm 0.00$ & $0.65 \pm 0.03$ & $0.54 \pm 0.03$ & $0.69 \pm 0.01$ \\
\hline
\end{tabular}%
}
\end{table}

In contrast, Qwen2.5-72B-Instruct and Llama3-XuanYuan3-70B-Chat excel in MRR and HR@K, indicating they are more effective at mimicking the user's realized, and often noisy, decision-making patterns. This gap reveals a fundamental trade-off: $y_{util}$ acts as a financially robust reference to align with the user's latent psychology, whereas $y_{user}$ captures the empirical behavior which may deviate from pure rationality. \textbf{The results suggest that models must navigate the tension between providing the most rational advice and maintaining empathetic behavioral alignment}.

% However, higher uNDCG does not necessarily translate into better recovery of the user’s realized choices. Metrics that reflect behavioral alignment, such as MRR and Hit Rate, exhibit substantially greater variance across models. Qwen2.5-72B-Instruct and Llama3-XuanYuan3-70B-Chat achieve the strongest performance on MRR and HR@K, with HR@3 reaching 0.69 in both cases, indicating a higher likelihood of placing the user-selected stock among top recommendations. In contrast, models such as GPT-5.2 and Qwen3-235B-A22B-Instruct show competitive uNDCG but comparatively lower HR@1 and HR@3, reflecting weaker alignment with individual decision outcomes despite strong preference-level ranking quality.

% Across models, HR@3 is substantially relatively robust than HR@1 and more consistent with uNDCG scores, and shows lower variance across models, indicating that models are less effective at precise choice performance (HR@1) and better at set-level accuracy of placing the user-selected stock within a small candidate set (HR@3).

% This gap highlights a key challenge of the task: aligning with a latent preference structure is easier than accurately predicting a single realized choice, especially under noisy, multi-objective decision dynamics.

\subsection{Expert Alignment Analysis}

Table~\ref{tab:eas} reveals how models resolve competing advisory principles. A prominent trend is the strong coupling between Rational Utility and Market Momentum alignment, which stems from the contextual collinearity of these signals during trending markets where high-momentum assets often dominate utility calculations. Llama-3.3-70B-Instruct exemplifies this trend by achieving the highest alignment with both Utility and Momentum, yet its sharp decline in Risk alignment proves that it struggles to decouple downside protection from growth-oriented signals.

% Table~\ref{tab:eas} further reveals systematic differences in how models resolve competing advisory principles. Llama-3.3-70B-Instruct exhibits substantially higher alignment with both the \textsc{Utility} ($\tau=0.74$) and \textsc{Momentum} ($\tau=0.73$) experts than all other models, suggesting a strong tendency to follow coherent, internally consistent ranking strategies. However, its relatively low alignment with the \textsc{Safety} expert indicates a bias toward return-driven or balanced strategies rather than conservative, downside-focused reasoning.

% Across all models, alignment with \textsc{Utility} and \textsc{Momentum} is consistently tracking one another, whereas alignment with \textsc{Safety} is systematically weaker, typically at roughly half the magnitude, highlighting a tendency to prioritize growth- or return-oriented reasoning over explicit risk aversion.

\begin{table}[t]
\centering
\caption{Alignment of model-generated rankings with heterogeneous advisory principles.}
\label{tab:eas}
\resizebox{\linewidth}{!}{%
\begin{tabular}{lccc}
\hline
Model & $\tau$(Utility)$\uparrow$ & $\tau$(Momentum)$\uparrow$ & $\tau$(Risk)$\uparrow$ \\
\hline
Random  & $0.00 \pm 0.01$ & $0.00 \pm 0.01$ & $0.00 \pm 0.01$ \\
GPT-5.2 & $0.59 \pm 0.02$ & $0.56 \pm 0.02$ & $0.28 \pm 0.02$ \\
GPT-4o  & $0.60 \pm 0.02$ & $0.60 \pm 0.02$ & $0.20 \pm 0.02$ \\
% Gemini-3 & -- & -- & -- \\
DeepSeek-V3.2 & $0.51 \pm 0.02$ & $0.49 \pm 0.02$ & $0.26 \pm 0.02$ \\
Qwen3-235B-A22B-Instruct & $0.56 \pm 0.02$ & $0.55 \pm 0.02$ & $0.26 \pm 0.02$ \\
Qwen2.5-72B-Instruct & $0.52 \pm 0.02$ & $0.49 \pm 0.02$ & $0.22 \pm 0.02$ \\
Llama-3.3-70B-Instruct & $0.74 \pm 0.02$ & $0.73 \pm 0.01$ & $0.17 \pm 0.02$ \\
Llama3-XuanYuan3-70B-Chat  & $0.47 \pm 0.02$ & $0.46 \pm 0.02$ & $0.15 \pm 0.02$ \\
\hline
\end{tabular}%
}
\end{table}

In contrast, DeepSeek-V3.2 demonstrates the most balanced profile across all evaluated models. By maintaining a stable and relatively high alignment with Safety while avoiding extreme bias toward return-driven metrics, DeepSeek-V3.2 shows a superior ability to integrate conflicting advisory signals into a compromise recommendation. The GPT series also exhibits similar balanced characteristics, though with slightly less consistency than DeepSeek in the safety dimension.

The behavior of Llama3-XuanYuan3-70B-Chat is particularly noteworthy given its background as a domain-specific LLM fine-tuned on financial corpora. Despite its lower expert alignment scores, it achieves high behavioral hit rates in Table~\ref{tab:overall-performance}, suggesting that its financial expertise manifests as empathetic alignment with User Choices rather than strict adherence to idealized mathematical formulas. XuanYuan3 acts as a seasoned human consultant who prioritizes the pragmatic, albeit noisy, preferences of real-world investors over rigid algorithmic consistency.

% In contrast, GPT-5.2 and GPT-4o demonstrate more balanced expert alignment profiles, with moderate Kendall’s $\tau$ across utility and momentum dimensions and non-negligible alignment with safety. This suggests that these models integrate multiple advisory signals more evenly, rather than strongly favoring a single principle.

% Open-source models exhibit diverse behaviors: DeepSeek-V3.2 and Qwen3-235B-A22B-Instruct show moderate, relatively uniform alignment across all three experts, whereas Llama3-XuanYuan3-70B-Chat shows weaker overall alignment, consistent with its lower EAS scores despite strong behavioral metrics in Table~\ref{tab:overall-performance}.

\subsection{Preference Discovery Dynamics}

\begin{figure}[h]
  \centering
  \includegraphics[width=.95\linewidth]{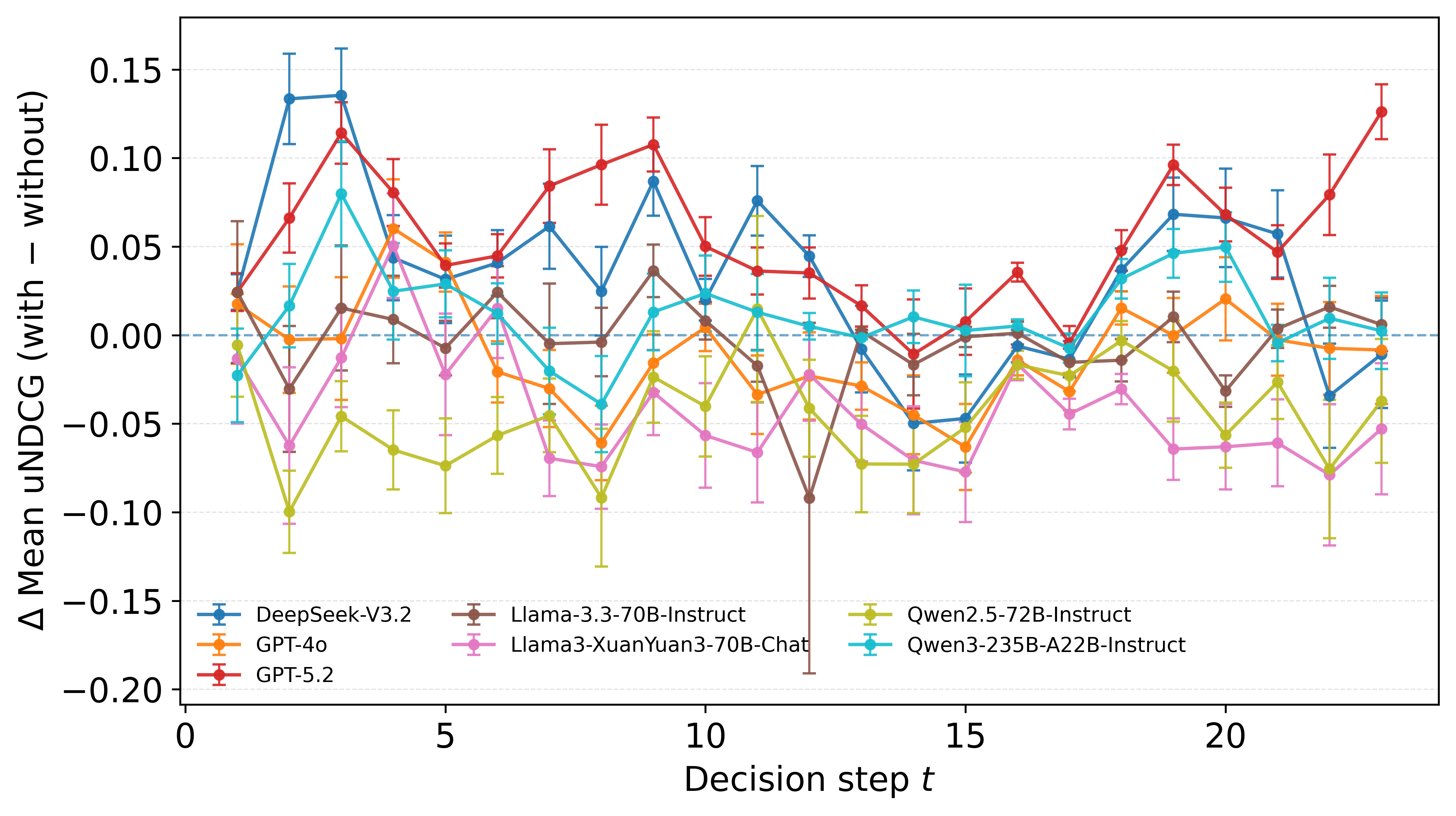}
  \caption{Step-wise improvement in utility-based alignment from conversational history.}\label{fig:dna_curve}
\end{figure}

Figure~\ref{fig:dna_curve} illustrates the step-wise gain ($\Delta$ uNDCG) in utility-based alignment when conversational history is accessible. While gains are observed for several models, the dynamics are highly heterogeneous. Models like GPT-5.2 and DeepSeek-V3.2 show significant positive improvements in early to middle stages (steps 1–10), suggesting they successfully extract informative signals about the user’s latent risk preferences from initial interactions. The subsequent fluctuations and general plateauing across most models indicate that while LLMs can form a coarse preference representation early on, the inherent "noise" in longitudinal financial decisions makes consistent, long-term preference tracking challenging. 

From a financial advisory perspective, these gains suggest that conversation history allows models to gauge baseline investment styles, identifying whether a user is inherently risk-averse or return-sensitive. However, the plateauing suggests that the marginal utility of additional historical context diminishes once a stable "investor persona" is established, at which point ranking performance becomes more contingent on the immediate market context $\mathcal{M}_t$ than on further preference refinement.

\begin{figure}[h]
  \centering
  \includegraphics[width=.9\linewidth]{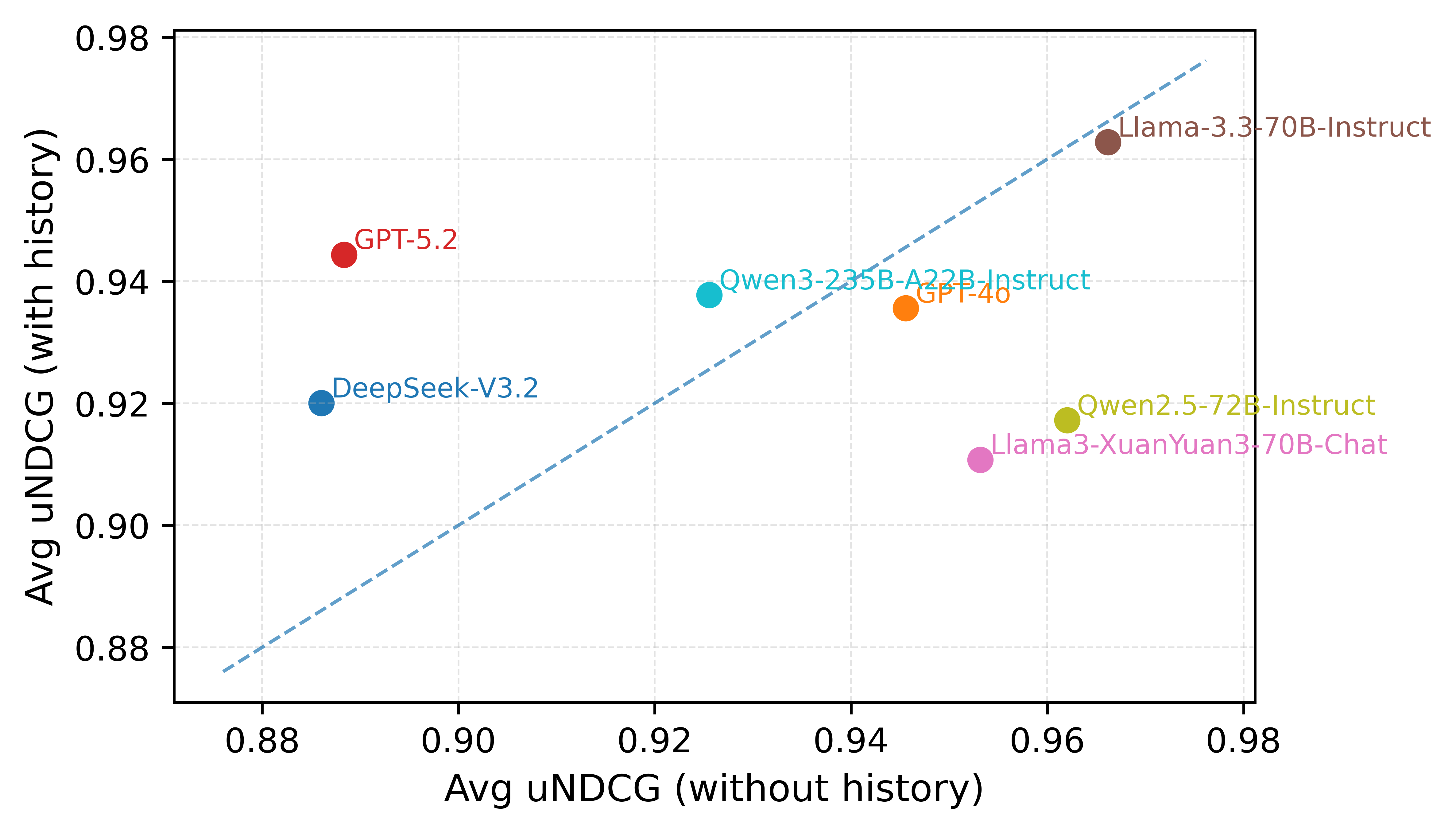}
  \caption{Average utility alignment with and without conversational history.}\label{fig:dna_scatter}
\end{figure}

Figure~\ref{fig:dna_scatter} provides a finer-grained diagnostic by comparing average utility alignment with and without longitudinal context. Based on the relative shifts from the diagonal, we identify three archetypes.

\textbf{Adaptive Advisors} (GPT-5.2, DeepSeek-V3.2, Qwen3-235B) show clear improvements when history is available, indicating effective cross-turn preference integration and progressive alignment with the user’s latent risk profile.

\textbf{Transaction-driven Analysts} (GPT-4o, Llama-3.3-70B) remain close to the diagonal, achieving strong utility rankings but exhibiting limited gains from conversational context, suggesting reliance on contemporaneous market signals rather than personalization.

\textbf{Behavioral Overfitters} (Qwen2.5-72B, Llama3-XuanYuan3) experience degraded utility alignment when history is introduced, implying over-sensitivity to noisy user actions and a tendency to prioritize behavioral mimicry over stable preference inference.

The results reveal non-uniform preference discovery and highlight the need to separate surface imitation from genuine decision-utility alignment, as evidenced by XuanYuan’s performance drop.

% Together, these results highlight that preference discovery is not a uniform capability. The observed performance drop in specialized models like XuanYuan emphasizes the need for evaluation protocols that distinguish between surface-level behavioral imitation and true decision-utility alignment.

\section{Conclusion}

We introduce \textsc{Conv-FinRe}, a conversational and longitudinal benchmark that shifts financial recommendation from surface-level behavioral matching to utility-grounded decision alignment. Through inverse optimization of latent risk preferences, it supports multi-view evaluation that separates rational decision quality from observed user behavior. Our results reveal a persistent tension between utility-based ranking and behavioral alignment: general-purpose LLMs often optimize utility more effectively, while domain-specific models tend to overfit transient user actions. These findings expose the limits of behavior-only evaluation and motivate benchmarks that disentangle long-term investor preferences from short-term market noise.

\bibliographystyle{ACM-Reference-Format}
\bibliography{sample-base}

% \appendix









\end{document}